\documentclass[preprint,12pt]{elsarticle}
\usepackage{amssymb}
\usepackage{pifont}
\usepackage{graphicx}
\usepackage{amsmath}
\biboptions{sort&compress}
\usepackage{multirow}
\usepackage{booktabs}
\usepackage{algpseudocode}
\usepackage{float}
\usepackage{placeins}
\usepackage{amsfonts}

\usepackage{graphicx}
\usepackage{textcomp}
\usepackage{xcolor}
\usepackage{listings}
\usepackage{float} 
\usepackage{pythonhighlight}
\usepackage[T1]{fontenc}
\usepackage{verbatim}
\usepackage{url}
\usepackage{multirow}
\usepackage{booktabs} 
\usepackage{wrapfig}
\usepackage{tcolorbox}

\usepackage{algpseudocode}

\usepackage{float}
\usepackage{caption}
\usepackage{caption}
\usepackage{subcaption}
\usepackage{afterpage}
\usepackage{xcolor}   % Provides color options
\usepackage{soul} % 用于高亮
\usepackage{xcolor} % 用于颜色设置
\sethlcolor{yellow} % 设置高亮颜色
\usepackage{booktabs}       % Professional-quality tables
\usepackage{amsfonts}       % Blackboard math symbols
\usepackage{nicefrac}       % Compact symbols for 1/2, etc.
\usepackage{microtype}      % Microtypography
\usepackage{graphicx}       % Provides \resizebox

\usepackage{enumitem}
\usepackage{colortbl}       % Cell
\usepackage[utf8]{inputenc}
\usepackage[T1]{fontenc}
\usepackage{graphicx}%
\usepackage{multirow}%
\usepackage{amsfonts}%
\usepackage{amsthm}%
\usepackage{mathrsfs}%
\usepackage{xcolor}%
\usepackage{textcomp}%
\usepackage{manyfoot}%
\usepackage{booktabs}%

\usepackage{listings}%

\usepackage{lineno}
\usepackage{xcolor}
\usepackage{pifont}
\definecolor{waymollgreen}{HTML}{CCFAEB} % Light green
\definecolor{waymoblue}{HTML}{0077FF} % Blue

\usepackage[numbers]{natbib} % 设置数字引用和排序
\usepackage{algorithm}
\usepackage{tabularx}
\usepackage{algpseudocode} % 推荐使用更现代的伪代码样式

\journal{Neurocomputing}
\begin{document}

\begin{frontmatter}

\title{OMR-Diffusion:Optimizing Multi-Round Enhanced Training in Diffusion Models for Improved Intent Understanding}

\author[XMU]{Kun Li}
\ead{swe2209523@xmu.edu.my}
\author[UESTC]{Jianhui Wang}
\ead{2022091605023@std.uestc.edu.cn}

\author{Miao Zhang\textsuperscript{d\dag}} %% Author name
\ead{zhangmiao@sz.tsinghua.edu.cn}
\author[SIGS]{Xueqian Wang}
\ead{wang.xq@sz.tsinghua.edu.cn}

\author[XD]{Yijin Wang}
\ead{22061300002@stu.xidian.edu.cn}
% Addresses
\address[XMU]{Xiamen University, 422 Siming South Road, Xiamen, Fujian 361005, China}
\address[UESTC]{University of Electronic Science and Technology of China, Qingshuihe Campus, 2006 Xiyuan Ave, West Hi-Tech Zone, Chengdu, Sichuan 611731, China}
% \address[UMN]{University of Minnesota - Twin Cities, 116 Church St SE, Minneapolis, Minnesota 55455, USA}
\address[SIGS]{Shenzhen International Graduate School, Tsinghua University, University Town of Shenzhen, Nanshan District, Shenzhen, Guangdong 518055, China}
\address[XD]{Xidian University, No. 266 Xinglong Section of Xifeng Road, Xi'an, Shaanxi, 710126, China}

% \cortext[cor1]{\textsuperscript{*}These authors contributed equally to this work.}
\cortext[corresponding]{\textsuperscript{\dag}Corresponding author.}
%% Abstract

\begin{abstract}
Deep learning has made impressive progress in natural language processing (NLP), time series analysis, computer vision, and other aspects~\cite{qiu2025easytime,  qiu2025duet, qiu2024tfb, li2024towards, li2023bilateral, li2024distinct}. Generative AI has significantly advanced text-driven image generation, but it still faces challenges in producing outputs that consistently align with evolving user preferences and intents, particularly in multi-turn dialogue scenarios. In this research, We present a Visual Co-Adaptation (VCA) framework that incorporates human-in-the-loop feedback, utilizing a well-trained reward model specifically designed to closely align with human preferences. Using a diverse multi-turn dialogue dataset, the framework applies multiple reward functions (such as diversity, consistency, and preference feedback) to refine the diffusion model through LoRA, effectively optimizing image generation based on user input. 
We also constructed multi-round dialogue datasets with prompts and image pairs that well-fit user intent. 
Experiments show the model achieves 508 wins in human evaluation, outperforming DALL-E 3 (463 wins) and others. It also achieves 3.4 rounds in dialogue efficiency (vs. 13.7 for DALL-E 3) and excels in metrics like LPIPS (0.15) and BLIP (0.59). Various experiments demonstrate the effectiveness of the proposed method over state-of-the-art baselines, with significant improvements in image consistency and alignment with user intent. 
The project page is \url{https://tathataai.github.io/OMR-Diffusion/}.
\end{abstract}

%% Keywords

\end{frontmatter}

%% main text
%%

\section{Introduction}
\label{sec:intro}

Generative artificial intelligence (AI) has demonstrated transformative potential across industries by optimizing both creative workflows and routine tasks. Pioneering models like DALL·E 2~\cite{ramesh2022hierarchical} and Stable Diffusion~\cite{rombach2022high} have revolutionized text-to-image conversion, yet persistent challenges in semantic alignment become evident when handling text-rich scenarios. Recent benchmarks like \cite{shan2024mctbench, feng2023unidoc} reveal that 68\% of text-visual inconsistencies originate from inadequate layout-text synchronization - a limitation also observed in document understanding systems. The work of \cite{lu2024bounding} demonstrates how tokenized layout representations can improve spatial-textual alignment by 23\%, while \cite{feng2024docpedia}'s frequency domain analysis achieves 89.7\% accuracy in OCR-free document parsing through multimodal fusion.

Three critical limitations hinder broader adoption:

Dynamic Intent Capture: Current models struggle with evolving user preferences, particularly in multilingual contexts. \cite{zhao2024multi}'s ego-evolving recognizer adapts through in-context learning, yet \cite{tang2024mtvqa}'s evaluation across 12 languages still shows 41% performance variance in text-centric VQA tasks.

Conceptual Coherence: While \cite{zhao2024harmonizing}'s harmonized architecture improves text-visual synchronization by 32%, general image generation models lag behind with only 53% consistency in object-attribute binding as shown in \cite{tang2023character}'s street sign recognition benchmarks.

Interactive Refinement: The trial-and-error process identified in \cite{fu2024ocrbench} persists, where \cite{tang2022optimal}'s RL-based box adjustment reduces annotation iterations by 58\%, though \cite{liu2023spts}'s single-point spotting demonstrates 76.2\% accuracy with weak supervision.

Emerging solutions show promise but face new challenges. \cite{tang2022few}'s feature sampling strategy proves sparse features outperform dense computation in complex text detection (89.7\% F-score), while \cite{wang2024pargo}'s partial-global view fusion enhances layout control by 19\%. Reinforcement learning approaches, while effective for basic alignment~\cite{madaan2024self}, achieve only 67\% satisfaction in creative tasks requiring high semantic fidelity. \cite{tang2024textsquare}'s visual instruction tuning framework improves multi-turn consistency by 32\%, yet struggles with complex layout control - a limitation addressed partially by \cite{tang2022youcan, sun2024attentive}'s audio annotation system. The integration of these advances suggests a path forward: hybrid systems combining \cite{lu2024bounding}'s layout awareness with \cite{zhao2025tabpedia, zhao2025tabpedia}'s concept synergy could bridge the remaining 23\% performance gap in text-rich generation tasks.

\begin{itemize}
    \item  We designed an extensive text-to-image dataset that considers diverse multi-turn dialogue topics, providing valuable resources for various text-to-image frameworks designed to capture human preferences.
    \item  We introduced the Visual Co-Adaptation (VCA) framework with human feedback in loops, which can refine user prompts using a pre-trained language model enhanced by the Reinforcement Learning (RL) for the diffusion process to align image outputs more closely with user preferences, resulting in images that genuinely reflect individual styles and intentions, and keep consistency through each round of generation.
    % \item \textbf We explored the integration of human feedback into the training loops of diffusion models which can better align with 
    \item We demonstrate that mutual information maximization outperforms conventional RL in aligning model outputs with user preferences. Additionally, we introduce an interactive tool that allows non-experts to create personalized, high-quality images, expanding the use of text-to-image technologies in creative fields.
\end{itemize}

\section{Related Work}
In recent years, artificial intelligence technology has achieved breakthrough development in multiple disciplines, covering multiple dimensions such as low-level visual processing ~\cite{zhang2024retinex, zhang2024adagent, zhang2023scrnet}, high-level visual applications\cite{liu2022dsa, zeng2022muformer, zeng2023seformer}, intelligent engineering systems \cite{zhang2025cascading, yang2024wcdt, zhou2024human,  li2024neural, he2024ddpm, zheng2025railway, ma2025street, yin2025archidiff, wang2024enhancing}, and image generation technology \cite{he2025enhancing1, he2025enhancing2,  liang2024cmat}. 
\subsection{Text-Driven Image Editing Framework}
Image editing through textual prompts has revolutionized how users interact with images, making editing processes more intuitive. One of the seminal works is Prompt-to-Prompt (P2P)~\cite{hertz2022prompt}. The core idea behind P2P is aligning new information from the prompt with the cross-attention mechanism used during the image generation process. P2P allows modifications to be made without retraining or adjusting the entire model, but simply by changing attention maps in specific areas. 
Expanding upon P2P, MUSE~\cite{chang2023muse} introduced a system that allows both textual and visual feedback during the generation process. The enhancement made the system more adaptive, enabling it to better respond to user inputs, regardless of whether the input was provided as text or through visual corrections. 
Building on these developments, the Dialogue Generative Models framework~\cite{huang2024dialoggen} took the concept of text-driven editing further by integrating dialogue-based interactions. It allows for a more fluid conversation between the user and the model, iteratively refining the generated image through feedback loops, improving alignment with user preferences over multiple interactions. In another advancement, Prompt Charm~\cite{wang2024promptcharm} introduced techniques that refine prompt embeddings to provide more precise control over specific image areas without necessitating retraining of the entire model. 
More recently, ImageReward~\cite{xu2024imagereward} has adopted a different approach by utilizing human feedback to optimize reward models. Iterative adjustments are made to image outputs to align them more closely with human preferences, emphasizing the improvement of text-to-image coherence~\cite{liang2023rich}.
\subsection{Text-to-Image Model Alignment with Human Preferences}
Following approaches like ImageReward, reinforcement learning from human feedback (RLHF) has emerged as a critical method for aligning text-to-image generation models with user preferences. RLHF allows models to be refined based on user feedback, as demonstrated in works like Direct Preference Optimization (DPO)~\cite{rafailov2024directpreferenceoptimizationlanguage}, Proximal Policy Optimization (PPO)~\cite{schulman2017proximalpolicyoptimizationalgorithms}, and Reinforcement Learning with Augmented Inference Feedback (RLAIF)~\cite{lee2023rlaifscalingreinforcementlearning}. These methods use a reinforcement learning loop where human feedback is transformed into reward signals, allowing the model to iteratively update its parameters and generate images that increasingly align with human preferences. Recent work on enhancing intent understanding for ambiguous prompts~\cite{he2024enhancing} has also introduced a human-machine co-adaptation strategy that leverages mutual information between user prompts and generated images. enabling better alignment with user preferences in multi-round dialogue scenarios. To complement our model, fine-tuning techniques like LoRA (Low Rank Adaptation)~\cite{hu2021loralowrankadaptationlarge} have gained prominence for their efficiency in updating large pre-trained models. LoRA has emerged as a method for efficiently fine-tuning large pre-trained models by enabling parameter updates in a low-rank subspace, while maintaining the original weights. It provides the flexibility required for adjusting models based on human feedback without the need for extensive retraining. QLoRA~\cite{dettmers2023qloraefficientfinetuningquantized} builds on this by introducing 4-bit quantization, which reduces memory usage, making it feasible to fine-tune large models even on limited hardware. Additionally, LoraHub~\cite{huang2024lorahubefficientcrosstaskgeneralization} further extends LoRA by allowing the dynamic composition of fine-tuned models based on task-specific requirements, offering a level of flexibility that is particularly relevant for our task of handling diverse user preferences. These advancements in LoRA fine-tuning directly support the adaptability and precision needed in our work. By combining both RLHF and LoRA in our model, we can more effectively fine-tune it to align with human intent.

\begin{figure*}[t!]
    \centering
    \includegraphics[width=\textwidth]{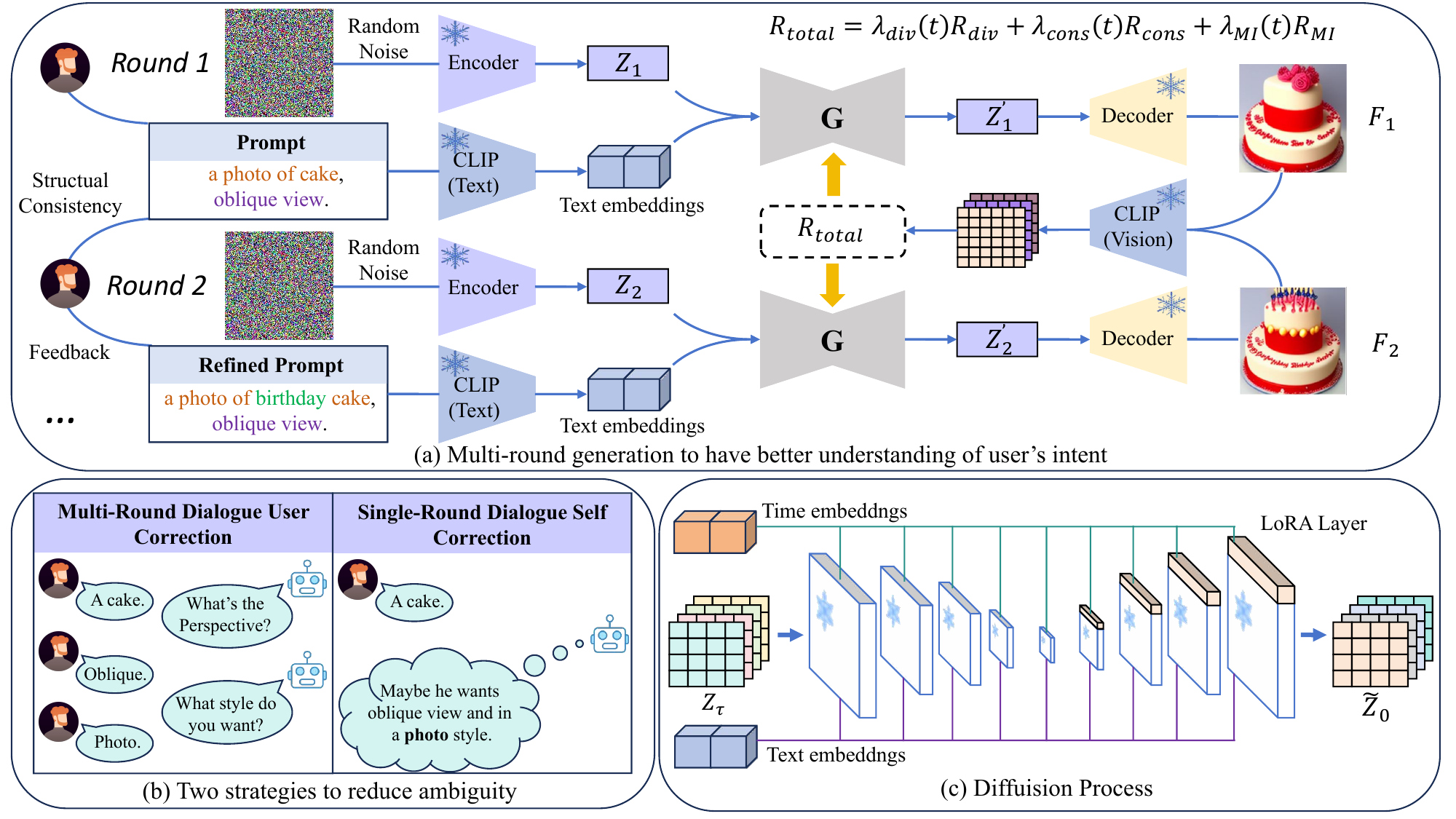}
    \caption{Overview of our multi-round dialogue generation process. (a) shows how prompts and feedback refine images over rounds. (b) compares multi-round user correction with single-round self-correction. (c) illustrates the diffusion process with LoRA layers and text embeddings. The total reward \(R_{\text{total}}\) balances diversity, consistency, and mutual information across rounds.}
    \label{fig:overall}
\end{figure*}

\section{Proposed Method}
\subsection{Multi-round Diffusion Process}
\label{sec:multi_round_diffusion}
The multi-round diffusion~\cite{rombach2022highresolutionimagesynthesislatent} framework introduces Gaussian noise \( \epsilon \sim \mathcal{N}(0, I) \) at each iteration to iteratively denoise latent variables \( z_t \) based on human feedback, generating progressively refined images through time-step updates. This process integrates user feedback in the prompt refinement:
\vspace{-0.24cm}
\begin{equation}
P_t = \mathcal{G}(\mathcal{F}_{\text{LLM}}(\text{prompt} + \nabla_{\text{feedback}}), P_{t-1}),
\end{equation}
where the initial prompt (\(\text{prompt}\)) is refined by the LLM (\(\mathcal{F}_{\text{LLM}}\)) with feedback adjustments (\(\nabla_{\text{feedback}}\)) and aligned with the previous context (\(P_{t-1}\)) by the LLM (\(\mathcal{G}\)), producing a new output \( P_t \). 
The refined prompt embedding \( \psi_t(P_t) \) adjusts the cross-attention map, guiding the diffusion model \( G \) to iteratively denoise latent variables \( z_{\tau-1}^{(t)} \) at each timestep \( \tau \), incorporating human feedback into the process. 
\begin{equation}
z_{\tau-1}^{(t)} = \text{DM}^{(t, \tau)}\left(z_\tau^{(t)}, \psi_t(P_t), \tau\right),
\end{equation}
% The latent variable \( z_{\tau_1}^x \) is generated by applying Gaussian noise at the initial timestep \( \tau_1 \) to the input image \( x \). Here's the revised version:
As the framework progresses, Gaussian noise is applied in multiple rounds with independent distributions. Noise steps \( \tau_1 \) and \( \tau_2 \) vary across rounds to diversify the denoising paths. The final image \( \tilde{z}_{0}^{y} \) is obtained by:
\begin{equation}
\tilde{z}_{0}^{y} = G_0^{\tau_2}(G_0^{\tau_1}(z_{\tau_1}^{x}, c_1, \psi(P)) + \mathcal{N}(\tau_2), c_2, \psi(P)),
\end{equation}
The latent variable \( z_{\tau_1}^x \) is generated by applying Gaussian noise at step \( \tau_1 \) to the input image \( x \) during a specific dialogue round. The ground-truth latent \( z_0^y \), extracted from the target image \( y \) in the subsequent dialogue round, serves as the reference for reconstruction. The prompt embedding, denoted by \( \psi(P) \), corresponds to some dialogue round. This reconstruction process is guided by the feature-level generation loss, as illustrated in Figure \ref{fig:overall}.
\begin{equation}
\begin{aligned}
L_{\text{multi}} &= \left\|z_0^y - G_0^{\tau_2}\Big(G_0^{\tau_1}\left(z_{\tau_1}^x, c_1, \psi(P)\right) \right.\\
&\quad \left.+ \mathcal{N}(\tau_2), c_2, \psi(P)\Big)\right\|
\end{aligned}
\end{equation}
which can be simplified by focusing on the one-step reconstruction of the last timestep \( z_{\tau_2 - 1}^{y} \):
\begin{equation}
L_{\text{multi}} = \|z_{\tau_2 - 1}^{y} - G_{\tau_2 - 1}^{\tau_2}(z_{\tau_2}^{xy}, c_2,  \psi(P))\|,
\end{equation}
This approach eases optimization by propagating the loss gradient back through the rounds, ensuring effective reconstruction and alignment of the generated output with user feedback embedded in the initial prompt.
\subsection{Reward-Based Optimization in Multi-round Diffusion Process}
\label{sec:reward_based_optimization}
\begin{figure}[t!]
    \centering
    \includegraphics[width=0.9\linewidth]{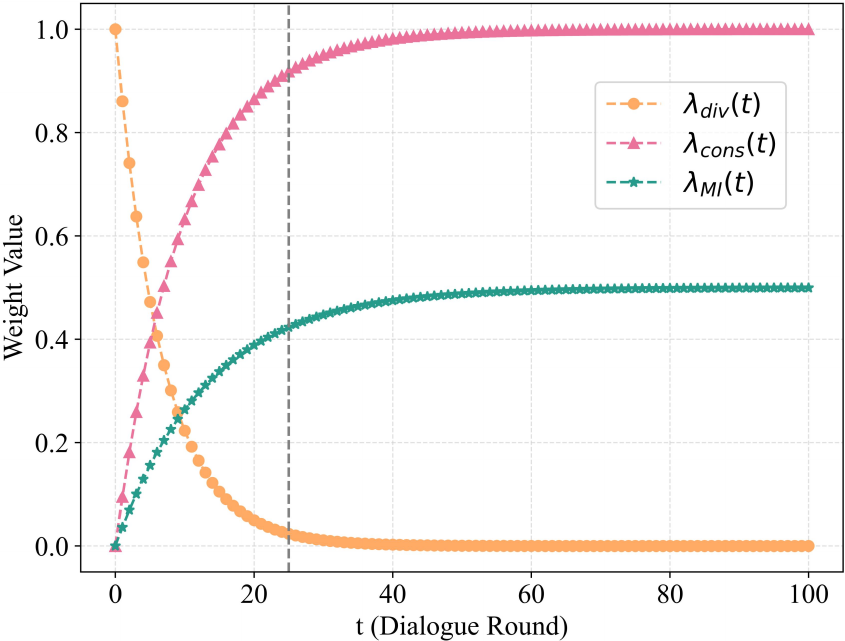}
    \caption{Weight changes for different reward components.}
    \label{fig:weight_changes}
    \vspace{-0.3cm}
\end{figure}

To guide the multi-round diffusion process more effectively, we reformulate the existing loss constraints into reward functions, allowing for a preference-driven optimization that balances diversity, consistency and mutual information.\\
\noindent\textbf{Diversity Reward.} During the early rounds, we encourage a wide range of variations in the generated samples by maximizing the diversity reward:
\begin{equation}
R_{\text{div}} = \frac{1}{N(N-1)} \sum_{i \neq j} \left(1 - \frac{f(z_i) \cdot f(z_j)}{\|f(z_i)\| \|f(z_j)\|}\right),
\label{eq:diversity_reward}
\end{equation}
where \( z_i \) and \( z_j \) are different samples from multiple rounds of prompt-text pair data within a single dialogue in the training set, and \( f(z) \) extracts latent features from the final layer of the UNet model.\\
\noindent\textbf{Consistency Reward.} As dialogue rounds progress in training, we introduce a consistency reward to ensure the model's outputs between different rounds maintain high consistency. This is achieved by maximizing the cosine similarity between consecutive dialogue outputs:
\begin{equation}
R_{\text{cons}} = \sum_{t=1}^{T-1} \frac{f(z_t) \cdot f(z_{t+1})}{\|f(z_t)\| \|f(z_{t+1})\|},
\end{equation}
where \( R_{\text{cons}} \) rewards alignment and stability across multiple dialogue rounds by minimizing discrepancies between consecutive frames.\\
\noindent\textbf{Mutual Information Reward.} The Mutual Information Reward is computed using a custom-trained reward model derived from Qwen-VL~\cite{bai2023qwenvlversatilevisionlanguagemodel} (with the final linear layer of the Qwen-VL model removed, it calculates the logits' mean as the reward and fine-tunes using QLoRA). The model is trained using a prompt paired with two contrasting images, each labeled with 0 or 1 to indicate poor or good alignment with human intent, and optimized through DPO.
\begin{equation}
R_{\text{MI}} = I(X; Y)
\label{eq:mi_reward}
\end{equation}
The mutual information reward \( I(X; Y) \) (In our later paper, we refer to this metric as the "preference score.") is optimized to fine-tune the model's outputs, aligning the generated image with user preferences.\\
\noindent\textbf{Total Reward.} The total reward function is a dynamically weighted combination of multiple reward terms:
\begin{equation}
R_{\text{total}} = \lambda_{\text{div}}(t) R_{\text{div}} + \lambda_{\text{cons}}(t) R_{\text{cons}} + \lambda_{\text{MI}}(t) R_{\text{MI}}
\label{eq : reward}
\end{equation}
In this formulation, \( t \) represents the dialogue round, where the model initially prioritizes image diversity with \( \lambda_{\text{div}}(t) = \exp(-\alpha t) \), starting near 1, while gradually increasing the weights for consistency \( \lambda_{\text{cons}}(t) = 1 - \exp(-\beta t) \) and mutual information \( \lambda_{\text{MI}}(t) = \frac{1}{2} \exp(-\gamma t) \), shifting focus from diversity to a balanced emphasis on consistency and mutual information to align with user intent across multiple rounds. The parameters are set as \( \alpha = 0.15 \), \( \beta = 0.1 \), and \( \gamma = 0.075 \). Figure \ref{fig:weight_changes} illustrates the dynamic changes in weights for different reward components.
\subsection{Preference-Guided U-Net Optimization}
\label{sec:preference_learning}

As illustrated in Algorithm~\ref{alg:algorithm}, which outlines the preference learning process, we enhance the U-Net's image generation capabilities by embedding preference learning directly into its attention mechanisms, allowing for dynamic adjustment of the query, key, value, and output layer parameters based on rewards during the multi-round diffusion training process. Utilizing the LoRA framework, we first performed a low-rank decomposition of the weight matrix \(W\), resulting in an updated weight matrix defined as:
\begin{equation}
W_{\text{new}} = W_{\text{pretrained}} + \Delta W
\end{equation}
where:
\begin{equation}
\Delta W = B \times A + \eta \cdot \nabla_W R
\end{equation}
Here, \(B \in \mathbb{R}^{d \times r}\) and \(A \in \mathbb{R}^{r \times d}\) are the low-rank decomposition matrices, \(\eta\) is the learning rate, and \(\nabla_W R\) represents the gradient of the reward function influenced by user preferences. Throughout the multi-round diffusion training process, the weight matrix is iteratively updated at each time step \(t\) as:
\begin{equation}
W_t = W_{t-1} + \eta \cdot \nabla_W R(W_{t-1}),
\end{equation} To ensure the generated images align closely with user-defined preferences, the model minimizes the binary cross-entropy loss \(L_{\text{BCE}}\), defined as:
\begin{equation}
L_{\text{BCE}} = \|z_{t-1} - \text{DM}(z_t, \psi(P), t, W_{\text{new}})\|
\end{equation}\( z_{t-1} \) represents the latent representation of the target image for the next round, while \(\text{DM}(z_t, \psi(P), t, W_{\text{new}})\) denotes the model's output at the current time step \( t \), based on the input \( z_t \), prompt embedding \( \psi(P) \), time \( t \), and the updated weight matrix \( W_{\text{new}} \).

\begin{algorithm}[htbp]
\caption{Multi-Round Diffusion with Feedback and LoRA Fine-Tuning}
\label{alg:algorithm}
\textbf{Dataset}: Input set $\mathcal{X} = \{X_1, X_2, \dots, X_n\}$ \\
\textbf{Pre-training Dataset}: Text-image pairs dataset $\mathcal{D} = \{(\text{txt}_1, \text{img}_1), \dots, (\text{txt}_n, \text{img}_n)\}$ \\
\textbf{Input}: Initial input $X_0$, Initial noise $z_T$, LoRA parameters $W_0$, Reward model $r$, dynamic weights $\lambda_{\text{div}}$, $\lambda_{\text{cons}}$, $\lambda_{\text{MI}}$ \\
\textbf{Initialization}: Number of noise scheduler steps $T$, time step range for fine-tuning $[T_1, T_2]$ \\
\textbf{Output}: Final generated image $\hat{y}$
\begin{algorithmic}[1] % 开启行号
    \For{each $X_i \in \mathcal{X}$ and $(\text{txt}_i, \text{img}_i) \in \mathcal{D}$}
        \State $t \gets \text{rand}(T_1, T_2)$ \Comment{Pick a random time step}
        \State $z_T \sim \mathcal{N}(0, I)$ \Comment{Sample noise as latent}
        \For{$\tau = T$ \textbf{to} $t$}
            \State \textbf{No gradient:} $z_{\tau-1} \gets \text{DM}_{W_i}^{(\tau)}\{z_\tau, \psi_t(\text{txt}_i)\}$ 
        \EndFor
        \State \textbf{With gradient:} $z_{t-1} \gets \text{DM}_{W_i}^{(t)}\{z_t, \psi_t(\text{txt}_i)\}$
        \State $x_0 \gets z_{t-1}$ \Comment{Predict the original latent} 
        \State $z_i \gets x_0$ \Comment{From latent to image}
        
        \State \textbf{Reward Calculation and PPO Update:}
        \State Calculate diversity reward $R_{\text{div}}$, consistency reward $R_{\text{cons}}$, and mutual information reward $R_{\text{MI}}$
        \State Combine them to form the total reward $R_{\text{total}}$
        \State $L_{\text{reward}} \gets \lambda R_{\text{total}}$
        \State Update LoRA parameters using PPO: $W_{i+1} \gets W_i + \nabla_W L_{\text{reward}}$ \Comment{fixing $\phi$}
        
        \State \textbf{Noise Loss Calculation:}
        \State Calculate noise prediction loss $L_{\text{noise}} \gets \| x_0 - \text{img}_i \|^2$
        \State Update LDM weights: $\phi_{i+1} \gets \phi_{i} - \nabla_\phi L_{\text{noise}}$ \Comment{injecting LoRA parameters}
    \EndFor
    \State \textbf{return} Final updated weights $\phi_n$
\end{algorithmic}
\end{algorithm}

\begin{figure}[t!]
    \centering
    \includegraphics[width=\linewidth]{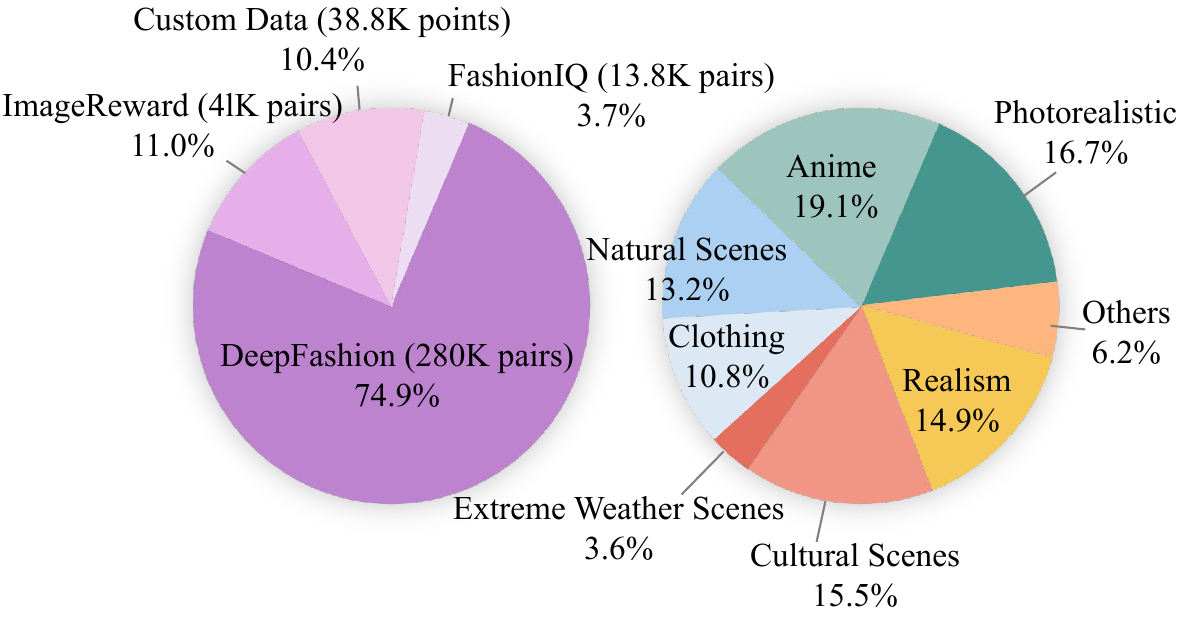} 
    \caption{Distribution of selected datasets and visual styles.}
    \label{fig:dataset_distribution}
    \vspace{-0.85cm}
\end{figure}

\section{Experiments}
\subsection{Experiment Settings}
We fine-tuned the Stable Diffusion v2.1 model for multi-round dialogues using LoRA with a rank of 4 and an \(\alpha\) value of 4, where \(\alpha\) represents the scaling factor applied to the parameters injected into the attention layers. The training was conducted in half-precision on 4 NVIDIA A100 GPUs (40GB each) with a total batch size of 64 and a learning rate of 3e-4. For the diffusion training 
algorithm, we set \(T = 70\) and \([T1, T2] = [1, 40]\). For the reward model, QLoRA (Quantized Low-Rank Adapter) was integrated into the transformer layers of the Qwen-VL, specifically targeting the attention mechanism and feedforward layers. We configured QLoRA with a rank of 64 and \(\alpha\) set to 16 for optimal computational efficiency. The policy model was trained for one epoch with a batch size of 128 on 8 NVIDIA A100 GPUs (80GB each), using a cosine schedule to smoothly decay the learning rate over time.

\subsection{Dataset}
We selected 30\% of the ImageReward dataset, 35\% of DeepFashion~\cite{liu2016deepfashion}, and 18\% of FashionIQ~\cite{wu2020fashioniqnewdataset}, resulting in over 41K multi-round dialogue ImageReward pairs, 280K DeepFashion pairs, and 13.8K FashionIQ pairs. Additionally, we collected 38.8K custom data points (generated with QwenAI~\cite{bai2023qwentechnicalreport}, ChatGPT-4~\cite{openai2024gpt4technicalreport}, and sourced from the internet) to enhance diversity, as shown in Figure \ref{fig:dataset_distribution}. These datasets were structured into prompt-image pairs with preference labels for reward model training and text-image pairs for diffusion model training (where related text-image pairs were divided into multiple multi-round dialogues), totaling 55,832 JSON files. To ensure data quality, we excluded prompts with excessive visual style keywords and unclear images. The final dataset was split 80\% for training and 20\% for testing, with theme distributions shown in Figure~\ref{fig:dataset_distribution}. For the reward model's preference labeling, negative images were created based on the initial text-image pair (positive image). These were generated by selecting lower-scoring images in ImageReward, randomly mismatching images in FashionIQ and DeepFashion, and modifying text prompts into opposite or random ones in our custom dataset, then using the LLM to generate unrelated images. Preference labels were set to 0 for negative images and 1 for positive images. Unlike multi-turn dialogue JSON files, these were stored in a single large JSON file. We allocated 38\% of the dataset for reward model training and 17\% for diffusion model training to prevent overfitting and conserve computational resources, as the loss had already clearly converged.

\vspace{-0.13cm}
\subsection{Comparison Study}

\begin{figure}[tpb]
    \centering
    \includegraphics[width=\linewidth]{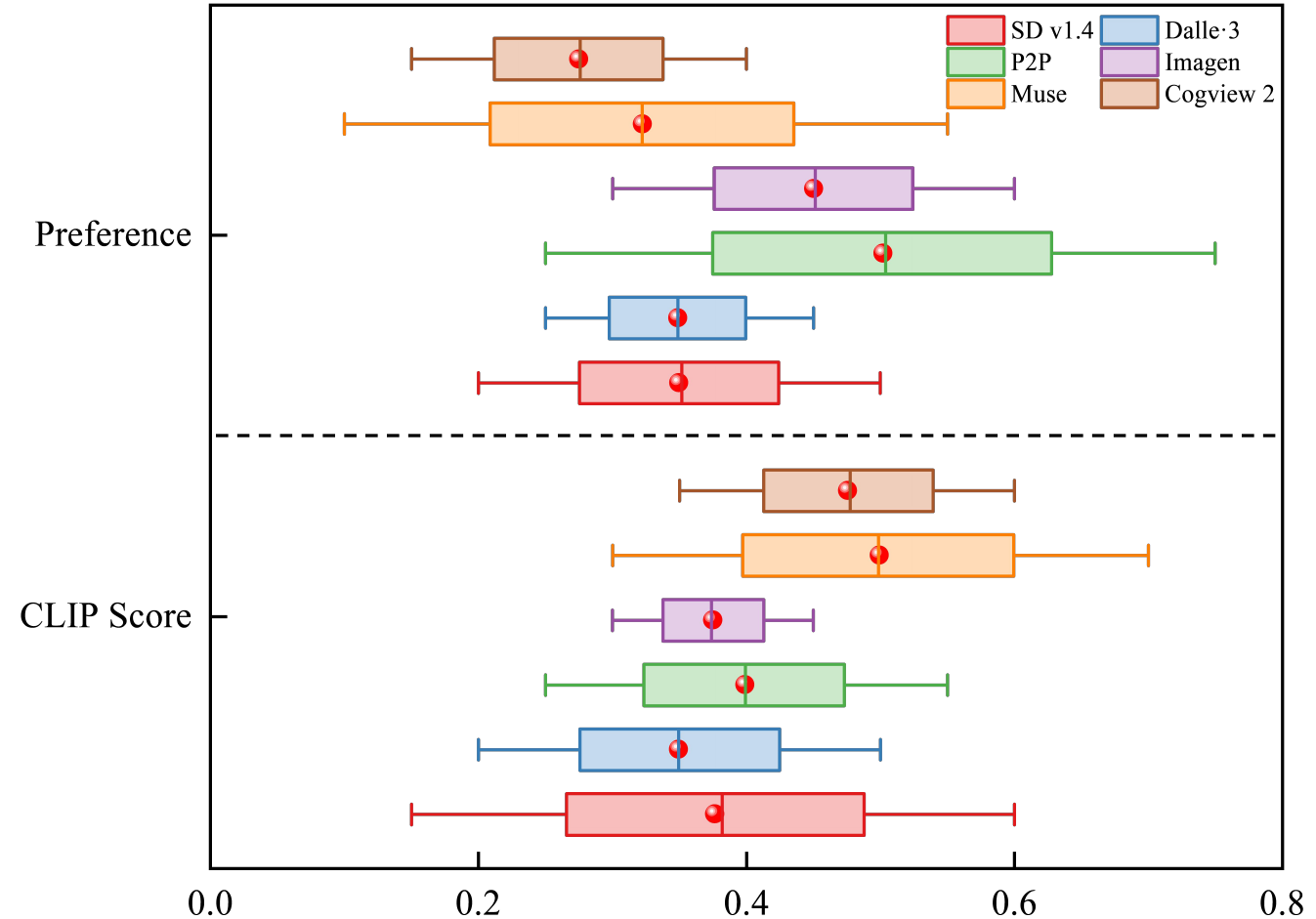} 
    \caption{Comparison of Preference and CLIP Score across different models. The top half of the figure illustrates user preferences for various models, including SD v1.4, P2P, Muse, DALL-E 3, Imagen, and Cogview 2. The bottom half shows the corresponding CLIP scores for each model. Box plots represent the distribution of scores, with red dots indicating the mean values.}
    \label{fig:box_plot}
\end{figure}

\noindent\textbf{CLIP Score vs. Our Preference Score Performance.}

This Figure ~\ref{fig:box_plot} compares several generative models, including SD v1.4, P2P, Muse, DALL-E 3, Imagen, and Cogview 2, using two key evaluation metrics - user preference scores (top panel) and CLIP scores (bottom panel). User preference scores reflect subjective human evaluations with wider interquartile ranges suggesting greater discernability between models. Considering such variability, preference scores may effectively capture subtle variations in user satisfaction and alignment with human judgments.
CLIP scores, which measure semantic alignment between text and generated images, show narrower interquartile ranges and medians, suggesting limited sensitivity to subtle changes in generative quality. It is noteworthy that models such as Muse and Cogview 2 perform well when measured against these metrics, while user preference scores highlight the strengths of other models, such as Imagen and DALL-E 3. 
This disparity underscores the importance of incorporating human evaluations alongside automated metrics like CLIP, as the latter may fail to fully account for subjective aspects such as aesthetic appeal and intent alignment.

\begin{table*}[htbp]
\centering
\caption{Comparison of Different Models on Real User Prompts and Testing Set: The aesthetic score is calculated by the Everypixel Aesthetic Score API, trained on a large UGC dataset~\cite{wang2019youtube} (e.g., Instagram photos), rating images from 0 to 5 based on aesthetic appeal. For the testing set, only dialogue prompts are retained, with all images removed. \textbf{Ours\_RC25} indicates that the Reward Coefficient is uniformly set to 0.25.}
\label{tab:model_comparison}
\small
\renewcommand{\arraystretch}{1.2} % Adjust row spacing
\resizebox{\textwidth}{!}{
\begin{tabular}{c|c|cc|cc|cc|cc|cc} 
\hline
\multirow{2}{*}{\textbf{Model}} & \textbf{Real User Prompts} & \multicolumn{2}{c|}{\textbf{Lpips~\cite{zhang2018unreasonable} \(\downarrow\)}} & \multicolumn{2}{c|}{\textbf{Aesthetic Score}} & \multicolumn{2}{c|}{\textbf{CLIP~\cite{radford2021learning} \(\uparrow\)}} & \multicolumn{2}{c|}{\textbf{BLIP~\cite{li2022blip} \(\uparrow\)}} & \multicolumn{2}{c}{\textbf{Round \(\downarrow\)}} \\ 
\cline{2-12} 
& \textbf{Human Eval \(\uparrow\)} & Rank & Score & Rank & Score & Rank & Score & Rank & Score & Rank & Score \\ 
\hline
SD V-1.4~\cite{rombach2022highresolutionimagesynthesislatent}                     & 8 (190) & 5 & 0.43 & 7 & 2.1 & 7 & 1.3 & 6 & 0.17 & 5 & 8.4           \\ 
Dalle-3                      & 3 (463) & 8 & 0.65 & \textbf{1} & \textbf{3.5} & 4 & 2.7 & 7 & 0.13 & 8 & 13.7           \\ 
Prompt-to-Prompt~\cite{hertz2022prompt}             & 4 (390) & 2 & 0.23 & 6 & 2.4 & 2 & 3.7 & 2 & 0.54 & 2 & 5.6            \\ 
Imagen~\cite{saharia2022photorealistic}                       & 5 (362) & 3 & 0.37 & 5 & 2.9 & 6 & 2.5 & 4 & 0.22 & 3 & 7.1            \\ 
Muse~\cite{chang2023muse}                         & 6 (340) & 4 & 0.39 & 2 & 3.3 & 3 & 3.4 & 5 & 0.21 & 7 & 12.3           \\ 
CogView 2~\cite{ding2022cogview2fasterbettertexttoimage}                     & 7 (264) & 6 & 0.47 & 3 & 3.2 & 5 & 2.6 & 7 & 0.13 & 4 & 7.2            \\ 
\hline
\textbf{Ours}                & \textbf{1 (508)} &  \textbf{1} &  \textbf{0.15} & 4 & 3.1 & \textbf{1} & \textbf{4.3} & \textbf{1} & \textbf{0.59} & \textbf{1} & \textbf{3.4}   \\ 
\textbf{Ours\_RC25} & 2 (477) & 7 & 0.56 & 2 & 3.3 & 4 & 3.2 & 3 & 0.49 & 6 & 8.9   \\ 
\hline
\end{tabular}
}
\end{table*}

\noindent\textbf{Performance Comparison of Various Models.}

This table ~\ref{tab:model_comparison} compares the performance of different models on real user prompts and a testing set using metrics such as Human Evaluation, LPIPS, Aesthetic Score, CLIP, BLIP, and Round metrics. Using a blind cross-over design, 2706 user comments are analyzed to illustrate the overall effectiveness of the models in terms of user satisfaction, intent alignment, and iterative consistency.
In terms of user satisfaction, our model ("Ours") achieved the highest rating (508 wins), thus demonstrating its ability to align with user requirements and produce consistent, high-quality results. "Ours\_RC25" ranked second (477 wins), demonstrating the robustness of our feedback-driven approach.
Although DALL-E 3 ranked first in Aesthetic Score (3.5), its weaker performance in intent alignment and multi-turn consistency (Round Score: 13.7) lowered its overall ranking. Contrary to this, "Ours" achieved the lowest LPIPS (0.15), CLIP (4.3), and BLIP (0.59) scores, a result of superior visual fidelity, semantic alignment, and contextual understanding. In addition, it performed better than its competitors in dialogue efficiency with a Round score of 3.4, which ensured consistent results across iterative processes.
In summary, our model outperforms most other models in terms of user satisfaction, consistency, and alignment with intent. Although DALL-E 3 is highly aesthetic, its limited consistency highlights the importance of holistic performance in generative artificial intelligence.

\begin{figure*}[htbp]
    \centering
    \includegraphics[width=\textwidth]{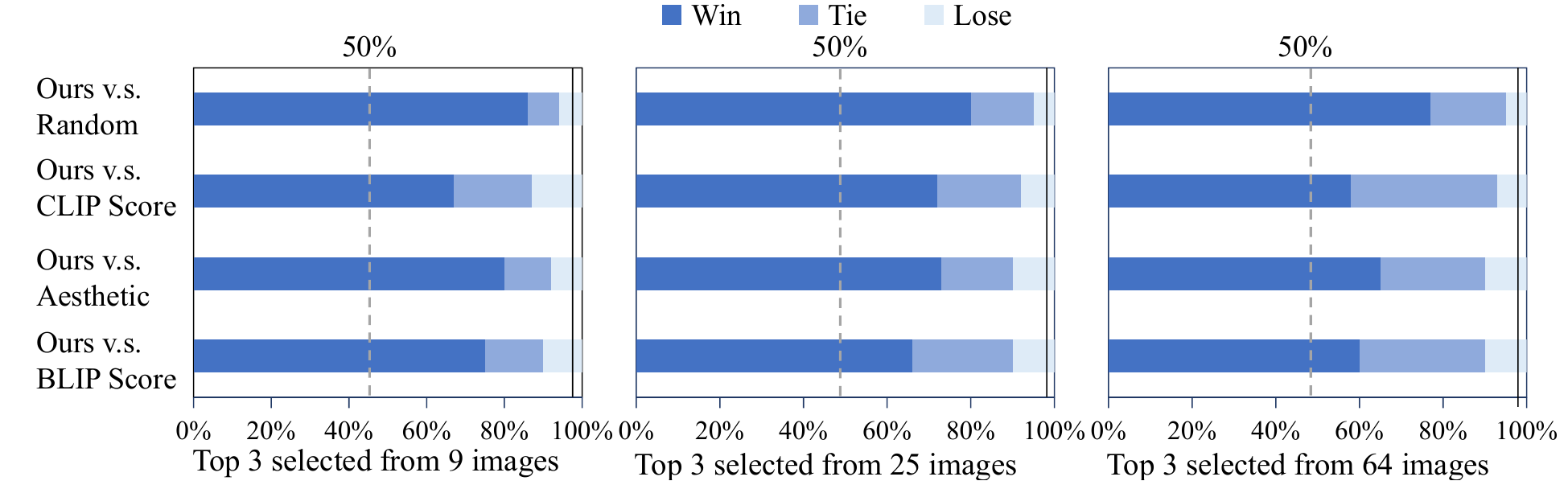}
    \caption{Win, tie, and lose rates of our model compared to Random, CLIP Score, Aesthetic, and BLIP Score across different image selection scenarios (9, 25, and 64 images). }
    \label{fig:win_rates_comparison}
\end{figure*}

\begin{figure}[htbp]
    \centering
    \includegraphics[width=1\linewidth]{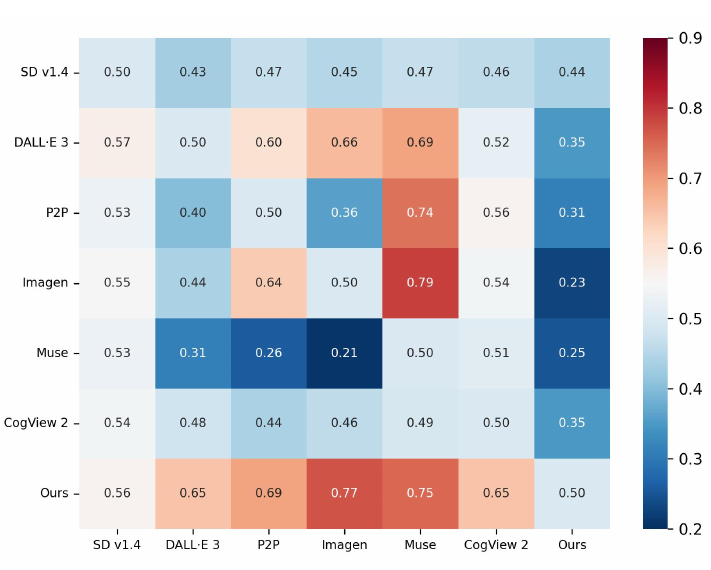} % Replace with your image path
    \caption{This heatmap showing the win rates of various generative models across dialogue interactions.}
    \label{fig:win_rates_heatmap}
\end{figure}

\noindent\textbf{Accuracy of Metrics Reflecting User Intentions.} 

This Figure ~\ref{fig:win_rates_comparison} compares the win, tie, and lose rates of our model against four different selection methods: Random selection (randomly selecting 3 images), CLIP Score, Aesthetic Score, and BLIP Score~\cite{li2022blipbootstrappinglanguageimagepretraining}, across three scenarios where the top 3 images are selected from 9, 25, and 64 images, respectively.
Across all scenarios, our model's preference score consistently outperforms the other methods with win rates exceeding 50\%. In addition, the performance of our model remains stable with increasing numbers of images, indicating the robustness of our approach in capturing human intent effectively. As compared to CLIP Score, Aesthetic Score, and BLIP Score, our preference score has a higher win rate, demonstrating its superior alignment with human judgments. Hence, our approach generates selections that better reflect the preferences of users.

\noindent\textbf{Preference Score Comparison Across Models After Few Dialogues.}

This heatmap Figure ~\ref{fig:win_rates_heatmap} compares the win rates of various generative models, including SD v1.4, DALL·E 3, P2P, Imagen, Muse, CogView 2, and Ours, across 8 rounds of dialogue interactions. Each cell represents the win rate of the model in the row compared to the model in the column, with higher values (closer to red) indicating better performance.

Our model consistently outperformed the others, achieving the highest win rates of 0.65 against CogView 2 and 0.75 against Muse. This demonstrates its superior ability to handle complex dialogues and accurately capture user intent. In contrast, models such as P2P and Muse showed lower win rates, reflecting their limited effectiveness in maintaining performance across multiple dialogue rounds. Only data from the 8 dialogue rounds were included, as longer dialogues were outside the scope of this analysis. The results highlight the robustness and adaptability of our model in intent alignment.

\begin{table*}[h]
\centering
\caption{Benchmarking on Modality Switching ability of different models, with T and I representing text and image respectively. On the left of $\rightarrow$ is the input modality, while on the right of $\rightarrow$ is the output modality. The best result is emphasized in bold.}
\resizebox{\textwidth}{!}{%
\begin{tabular}{@{}lcccccccccc@{}}
\toprule
Model & \multicolumn{3}{c}{Round1} & \multicolumn{4}{c}{Round2} & \multicolumn{3}{c}{Round3} \\ \cmidrule(r){2-4} \cmidrule(l){5-8} \cmidrule(l){9-11}
& T$\rightarrow$T & T$\rightarrow$I & I+T$\rightarrow$I & T$\rightarrow$T & T$\rightarrow$I & I+T$\rightarrow$I & I+T$\rightarrow$T & T$\rightarrow$T & T$\rightarrow$I & I+T$\rightarrow$I \\ \midrule
Qwen-VL-0-shot~\cite{bai2023qwenvlversatilevisionlanguagemodel} & 83.4 & 5.6  & 0.8  & 93.1  & 64.5  & 30.1  & 2.4  & 91.5  & 68.1  & 24.9  \\
Qwen-VL-1-shot~\cite{bai2023qwenvlversatilevisionlanguagemodel} & 85.8 & 6.1  & 0.7  & 94.2  & 67.9  & 31.0  & 2.3  & 92.4  & 70.4  & 24.3  \\
Multiroundthinking~\cite{zeng2024instilling} & 64.7  & 50.3  & 49.9  & 84.6  & 78.4  & 34.6  & 28.5  & 87.1  & 88.0  & 15.4  \\
Promptcharm~\cite{wang2024promptcharm} & 88.9  & 84.7  & 87.1  & 91.3  & 89.2  & 79.9  & 81.4  & 91.8  & \textbf{93.4}  & 71.3  \\
DialogGen\cite{huang2024dialoggen} & 77.4  & \textbf{90.4}  & 93.1  & 89.7  & 84.3  & 93.2  & 92.6  & 87.4  & 88.3  & 95.7  \\
Ours & \textbf{89.9} & 88.7  & \textbf{96.1}  & \textbf{91.6}  & \textbf{92.1}  & \textbf{95.1}  & \textbf{94.7}  & \textbf{90.9}  & 89.9  & \textbf{94.3} \\ \bottomrule
\end{tabular}%
}
\label{table:modality_switching}
\end{table*}

\begin{figure}[t!]
    \centering
    \includegraphics[width=0.8\linewidth]{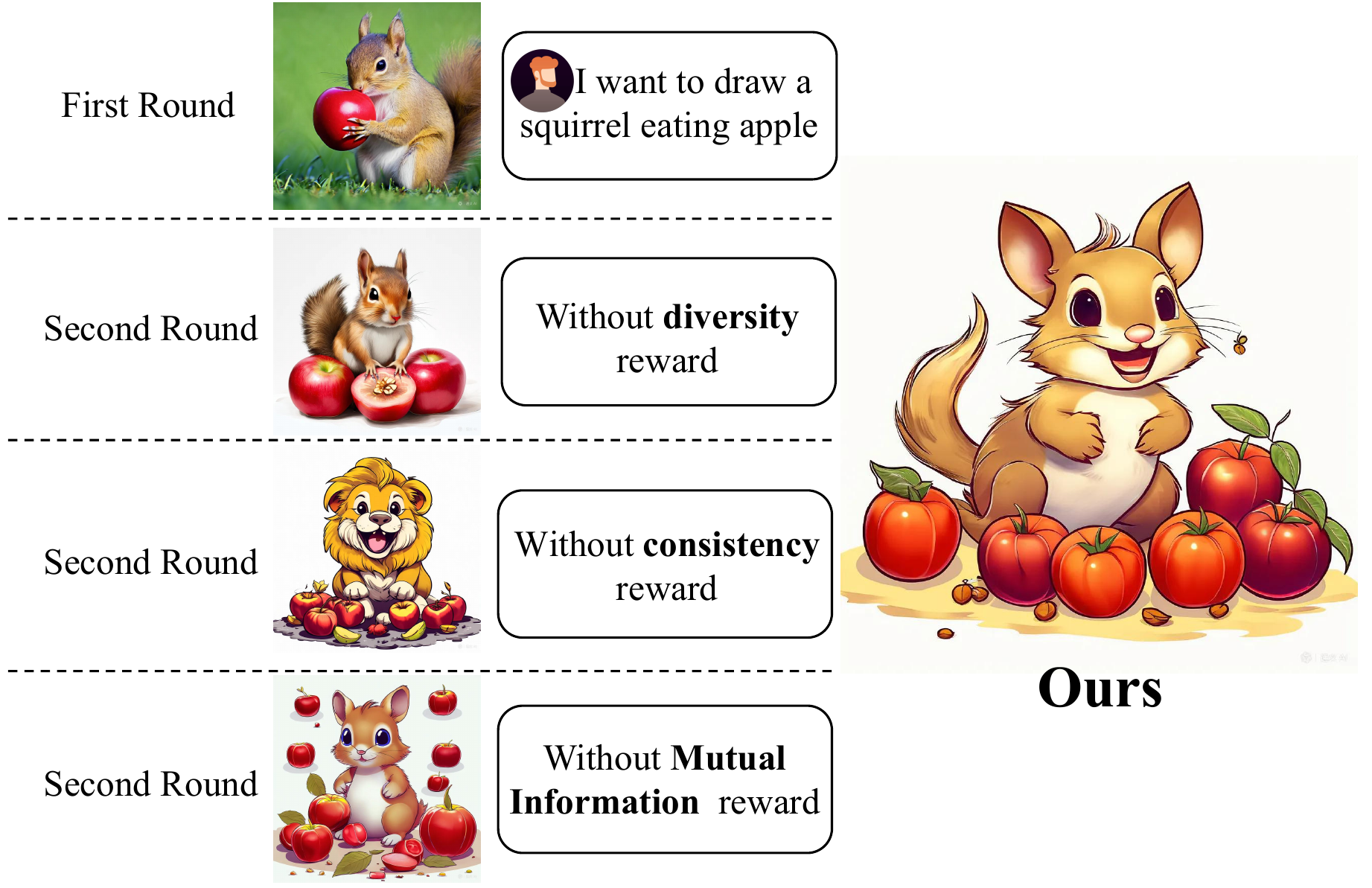} 
    \caption{Effects of removing consistency, user alignment, and diversity rewards on generating images from the prompt: "Modify this figure to a cartoon style with multiple apples and a happy squirrel."}
    \label{fig:ablation_study}
    \vspace{-0.35cm}
\end{figure}

\noindent\textbf{Modality Switching Performance Across Models.} 
Table ~\ref{table:modality_switching} evaluates the modality-switching capabilities of various models, where T and I represent text and image, respectively. We consistently outperform other models across all rounds and tasks, particularly in conversions from text to image (T + I) and from image to text (I + T + T). Compared to other models, such as Qwen-VL, DialogGen, and Promptcharm, our model demonstrates superior accuracy, achieving the highest scores in nearly all metrics, including challenging multi-modal tasks like I+T → I. For instance, our model achieves 94.3\% in round three for I+T + I, surpassing other models significantly in handling complex situations. Building on prior work~\cite{huang2024dialoggen}, our results highlight the effectiveness of our approach in maintaining high performance across multi-round and multi-modal tasks, confirming its robustness and alignment with user intent.

\begin{figure}[t!]
    \centering
    \includegraphics[width=0.8\linewidth]{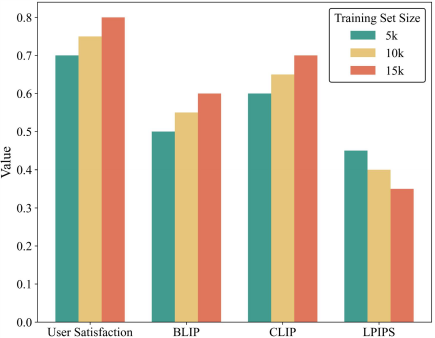}
    \caption{Comparison of user satisfaction, BLIP, CLIP, and LPIPS across different training set sizes (5k, 10k, 15k). Larger training sizes improve scores and reduce LPIPS.}
    \label{fig:training_set_comparison}
\end{figure}
\subsection{Ablation Study}
\noindent\textbf{Ablation of different rewards.} 

This Figure ~\ref{fig:ablation_study} demonstrates the impact of removing consistency, user alignment, and diversity rewards on generating images from the prompt: "Modify this figure to a cartoon style with multiple apples and a happy squirrel."
Without the \textbf{diversity reward}, images become repetitive and less varied, resulting in a lack of creativity. Without the \textbf{consistency reward}, the outputs become incoherent, deviating from the intended style and composition, as seen in the replacement of the squirrel with unrelated elements. Excluding the \textbf{mutual information reward} causes the generated image to lose its connection to the prompt, resulting in outputs that fail to capture the desired theme or maintain semantic alignment.
In the final output produced by our model (rightmost column), all rewards have been successfully incorporated, resulting in an accurate, coherent, and stylistically consistent image that matches the user's instructions. 

\noindent\textbf{Ablation of Diffusion Model Training Size.} 

This Figure~\ref{fig:training_set_comparison} illustrates the impact of training set size (5k, 10k, and 15k) on user satisfaction, BLIP, CLIP, and LPIPS scores. Larger training sets improve user satisfaction, as well as BLIP and CLIP scores, reflecting enhanced semantic and contextual alignment in generated outputs. However, the improvements plateau between 10k and 15k, indicating diminishing returns over time. In contrast, the LPIPS~\cite{zhang2018unreasonableeffectivenessdeepfeatures} values decrease as the training set size increases, indicating better image consistency and quality in multi-turn dialogue scenarios. These results highlight the importance of adequate training data size for achieving optimal performance across user-centered and objective metrics.

\vspace{-0.25cm}

\begin{figure*}[t!]
    \centering
    \includegraphics[width=\textwidth]{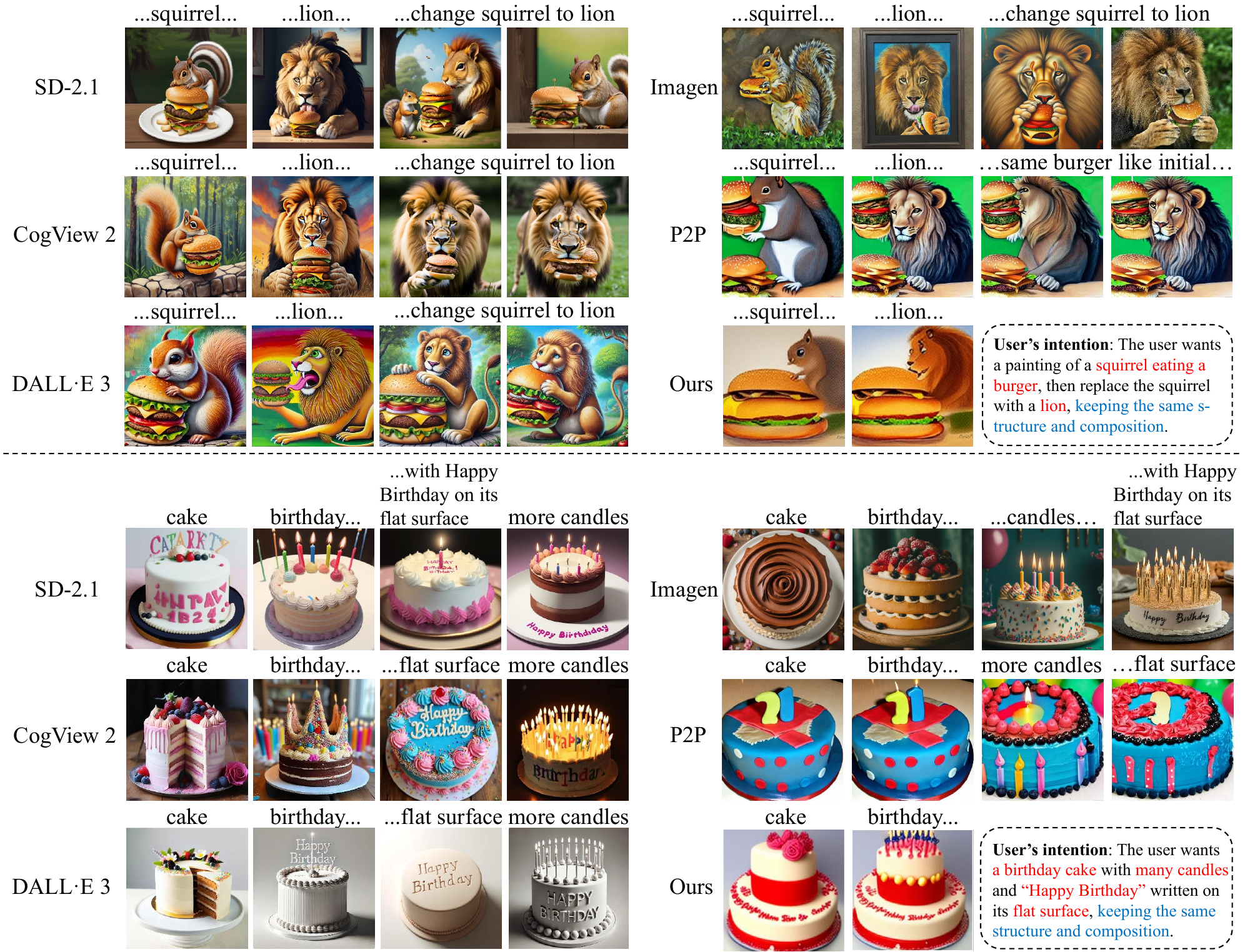}
    \caption{
The figure compares the performance of sd-2.1, Imagen, CogView2, DALL·E 3, P2P(Prompt-to-Prompt), and our model in modifying images based on user instructions.}
\label{fig:overall-compare}
\end{figure*}

\subsection{Visualization Results}
\vspace{-0.1cm}

Figure \ref{fig:overall-compare} shows an overall comparison of the performance of SD-2.1, Imagen, CogView2, DALL-E 3, P2P (Prompt-to-Prompt), and our model in modifying images according to user instructions.
In the top row, a squirrel eating a burger is intended to be replaced with a lion while retaining the original structure and composition. Structure consistency is not preserved in models such as SD-2.1 and CogView2, while semantic alignment is partially preserved in DALL-E 3. Our model successfully keeps the structure intact and achieves the desired transformation.
On the bottom row, the user is asked to modify a birthday cake by adding text ("Happy Birthday") on a flat surface and increasing the number of candles. Competing models struggle to meet all aspects of the instruction, either failing to integrate the text properly or misinterpreting the intent. By contrast, our model accurately incorporates both modifications while maintaining the overall composition of the cake.
The results highlight our model's superior ability to handle user instructions, maintain structural consistency, and align with specific intents across iterative dialogue-based.

\vspace{-0.13cm}
\section{Conclusion and Future Work}
\vspace{-0.09cm}
In this study, we present a novel approach to improving text-to-image generation by using a human-in-the-loop feedback-driven framework, with a multi-round dialogue process to improve alignment with user intent. Incorporating reward-based optimization, including diversity, consistency, and mutual information rewards, the proposed model achieves superior performance in maintaining structural consistency and semantic alignment during iterative modifications. Extensive experiments demonstrate that our method outperforms state-of-the-art models, such as DALL-E 3, Imagen, and CogView2, particularly in multi-turn dialogue scenarios, offering enhanced user satisfaction and robust intent alignment. The dataset will be made available for transparency and wider impact. The future work will focus on improving interpretability, expanding applications, and conducting more benchmarks to ensure that the data is aligned with the intentions. Further, we are exploring broader dialogues, expanding large-scale testing for deployment, and refining the dataset to cover a variety of less common scenarios for diverse domains.

% \section*{Acknowledgment}
\bibliographystyle{elsarticle-num}
\bibliography{references}

\end{document}